%% file: main.tex
\documentclass[sigconf,nonacm]{acmart}

\usepackage{bm}
\usepackage{algorithm}
\usepackage{algorithmic}

\theoremstyle{plain}
\newtheorem{theorem}{Theorem}
\newtheorem{proposition}{Proposition}
\newtheorem{corollary}{Corollary}
\theoremstyle{definition}
\newtheorem{definition}{Definition}

\newcommand{\pref}{\mathrm{pref}}
\newcommand{\E}{\mathbb{E}}

\providecommand{\WSDMRevisionRoot}{generated/wsdm27_revision_v3}

\title{Auditing Multi-Proxy Explanations of Web User Preferences:
Identification Boundaries and Abstention}

\author{Yu Zhang}
\orcid{0009-0003-5368-1356}
\affiliation{\institution{AMap Alibaba Group}\city{Beijing}\country{China}}
\email{yuanyu.zy@alibaba-inc.com}

\author{Zhihan Wang}
\orcid{0009-0000-8880-7172}
\affiliation{\institution{AMap Alibaba Group}\city{Beijing}\country{China}}
\email{yingchen.wzh@alibaba-inc.com}

\author{Yuhang Lu}
\orcid{0009-0006-6883-2227}
\affiliation{\institution{AMap Alibaba Group}\city{Beijing}\country{China}}
\email{lyh203611@alibaba-inc.com}

\author{Shuai Li}
\orcid{0009-0002-0986-2881}
\affiliation{\institution{AMap Alibaba Group}\city{Beijing}\country{China}}
\email{lion.lis@alibaba-inc.com}

\hypersetup{pdfauthor={Yu Zhang, Zhihan Wang, Yuhang Lu, Shuai Li}}

\begin{document}
\raggedbottom
\pagestyle{plain}

\begin{abstract}
Web systems infer user preferences from behavior, ratings, text, and
LLM-generated profiles, yet explaining one proxy may reveal its measurement
process rather than the preference of interest. We introduce an audit framework
that distinguishes when multi-proxy preference attributions are point
identified, partially identified, or should abstain relative to an explicit
reference distribution. Under feature-dependent, arbitrarily signed proxy bias,
a strict majority of uncontaminated proxy attributions lets their coordinatewise
median recover latent-attribution support and direction and, with externally
justified positive scales, magnitude. We prove the threshold is sharp over the
stated contamination class using response-valid indistinguishable worlds,
characterize sharp minimax error and exact compatible-target sets under
approximate cleanliness and scale uncertainty, and propagate clean-proxy
attribution error to local, global, and support conclusions. A frozen known-truth
benchmark spanning two response families, four proxy counts, every contamination
count, three bias strengths, and eleven methods maps the full boundary and
confirms the focal below-boundary comparison on twenty unseen seeds. Under
correct budgets, a direct attribution-tensor audit achieved $100\%$ coverage,
$95.40\%$--$97.05\%$ active-driver sign decisions, and no false signs. In a
temporally separated Amazon study, median aggregation reduced comparator-relative
conditional sign error for all four targets; six of eight prespecified
target--metric contrasts had Bonferroni intervals excluding zero, and a
controlled real-covariate perturbation favored the median on its prespecified
focal metric for all four targets. Compatible sets diagnosed that the current
proxy configuration rarely licensed strict-sign conclusions. Because comparator
calibration was weak, the Amazon evidence supports fitted-surface alignment
rather than latent-preference recovery.
\end{abstract}



\maketitle

\input{sections/01_introduction}

\input{sections/02_related_work}
\input{sections/03_problem}
\input{sections/04_method}
\input{sections/05_theory}
\input{sections/06_experiments}
\input{sections/07_conclusion}
\raggedbottom
\input{sections/08_ethical_considerations}

\vfuzz=2pt
\bibliographystyle{ACM-Reference-Format}
\bibliography{references}

\end{document}

%% file: sections/01_introduction.tex
\section{Introduction}

Modern personalization relies on preference profiles emitted by large language
models (LLMs) and behavioral systems: ``this user prefers Jazz.'' When a
stakeholder asks \emph{why}, explaining the profile model is natural, but the
result concerns an \emph{observed proxy}---a future behavior label, an explicit
rating, or an LLM-generated tag---rather than the latent preference itself.
Feature-dependent proxy construction can therefore create plausible but
misdirected explanations: price sensitivity may shape purchases, and review
style may shape an LLM label, even when neither reflects the shared construct.

\paragraph{A concrete example.} A behavioral proxy may attribute ``prefers
Jazz'' to \emph{average price}, while an LLM proxy attributes it to \emph{rating
tone}. Each feature may describe how that proxy is produced rather than what the
proxies share. Treating either attribution as ground truth can target an artifact
of measurement. The useful object is therefore the attribution common to enough
differently constructed measurements, together with a boundary beyond which that
common signal is not recoverable without additional information.

\paragraph{An attribution-level contamination model.} Relative to a declared
reference distribution, Shapley linearity decomposes proxy $k$'s attribution as
$\phi_{ki}(x)=a_k\theta_i(x)+\psi_i^{b_k}(x)$: a positively scaled latent
attribution plus proxy-specific contamination. No single proxy can separate
these terms without assumptions. After externally anchored scale calibration,
a strict majority of uncontaminated entries is sufficient, and---over an
arbitrary coordinatewise contamination class---necessary for uniform recovery.
We turn the scalar median's $50\%$ breakdown principle into a response-constrained
identification result, using response-valid constructions to establish the exact
boundary for an unobserved attribution induced by multiple proxy response functions:
point recovery below the boundary, a matching impossibility result at and
beyond it, and partial identification under approximate cleanliness and scale
uncertainty.

\paragraph{Contributions.}
\begin{itemize}
\item \textbf{Estimand.} We distinguish local signed
$\theta_i(x)$ from global $I_i=\E|\theta_i(X)|$ and derive the scaled
multi-proxy measurement model (Sec.~\ref{sec:problem}).
\item \textbf{Identification.} We prove strict-majority
recovery and a matching impossibility result realized by valid response
functions, derive sharp minimax error under approximate cleanliness, exact
pointwise compatible-target sets under scale uncertainty, and conditional
finite-sample recovery guarantees
(Sec.~\ref{sec:theory}).
\item \textbf{Evidence.}
A known-truth benchmark spans two response families, proxy counts 3, 4, 5, and
7, every contamination count, three strengths, and eleven estimators, with a
separate 20-seed confirmation at one below-boundary cell. A direct attribution-
tensor audit verifies compatible-set coverage and sign decisions under correct
budgets. On temporally separated Amazon data, we evaluate alignment to fixed
future-response surfaces, measurement-readiness decisions, and a single-regime
real-covariate mechanism check (Sec.~\ref{sec:experiments}).
\end{itemize}

Together, these results make the identification boundary operational for
explanations of proxy response functions. The distinguishing elements are
response-valid indistinguishable worlds yielding a matching lower bound, exact
compatible-target sets under joint contamination and scale uncertainty, and a
unified error interface that propagates proxy-attribution error to local,
global, and support conclusions.

%% file: sections/02_related_work.tex
\section{Related Work}

\paragraph{Proxy outcomes, weak supervision, and noisy annotators.}
Optimizing an observable proxy can reproduce systematic measurement bias when
the proxy is not the construct of interest~\citep{obermeyer2019dissecting}.
Weak-supervision and crowdsourcing methods instead combine noisy labeling
functions or annotators to estimate a latent label, commonly through a
generative confusion model~\citep{dawid1979maximum,ratner2016data}. Recent work
also studies identifiability with instance-dependent annotator corruption
\citep{nguyen2024crowd}. These methods aggregate \emph{labels} before fitting a
predictor. Our question is different but complementary: given proxy-specific
response functions, which parts of their \emph{local attribution vectors} can be
shared latent signal rather than differential proxy bias? A label-first
majority model is therefore an objective-shifted alternative in our controlled
known-truth comparison, not a matched baseline for the same attribution tensor
or a claim that attribution-level aggregation replaces weak supervision.

\paragraph{Aggregation across explanation methods.}
Many explanation ensembles aggregate methods or variants for one observed
model: saliency averaging~\citep{rieger2019aggregating}, featurewise $\ell_1$
centroids and aggregate Shapley values~\citep{bhatt2020evaluating}, median
smoothing~\citep{gan2022stable}, or optimized convex combinations
\citep{decker2024provably}. EnsembleSHAP certifies a random-subspace ensemble
\citep{wang2026ensembleshap}; Rashomon analysis instead retains partial-order
statements shared across good-fitting predictors~\citep{laberge2023rashomon}.
Method disagreement for a fixed prediction is documented
\citep{krishna2024disagreement}; these works address observed targets. We keep attribution
semantics fixed but aggregate response functions from proxies of an unobserved
target, studying measurement contamination, impossibility, and compatible
targets; within-proxy ensembling is orthogonal.
Separately, aggregate SHAP on observed support need not certify that a feature
is absent from the underlying function because off-support evaluations matter
\citep{bhattacharjee2025discard}. Accordingly, our recovered support is the
support of the declared attribution estimand, not a claim about functional
dependence of $q$.
Latent SHAP instead explains an observed model in a human-interpretable latent
\emph{feature representation}~\citep{malach2025latent}; our ``latent'' object is
an unobserved response target whose attribution must itself be identified.

\paragraph{Robust location under corrupted updates.}
The median is a classical robust location estimator~\citep{huber1964robust}.
Coordinatewise medians and trimmed means also aggregate worker gradients with
arbitrary worker failures and attain statistical rates~\citep{yin2018byzantine}.
Block-coordinate geometric
medians also have a $1/2$ breakdown point~\citep{acharya2022robust}.
Crucially, a coordinatewise minority guarantee need not preserve a useful
high-dimensional update direction: constructed examples based on inner
products can invalidate median SGD~\citep{xie2020fall}. These works estimate an
observable location or gradient under corrupted worker messages. We instead ask
whether an \emph{unobserved attribution} is identified from different proxy
response functions: positive proxy scales require external anchors, clean sets
may change with $(i,x)$, signed local estimates precede global absolute
importance, and our lower bound is realized by observationally identical valid
response functions. Building on this robust-location mechanism, we introduce a
latent-attribution estimand, prove a sharp identification boundary with a
matching impossibility construction realized by valid response functions, and
derive compatible-target sets under proxy-specific contamination.

\paragraph{Measurement error and structured recovery.}
Measurement-error and nonlinear-factor models identify latent quantities using
auxiliary instruments or multiple indicators under model-specific independence
and normalization restrictions~\citep{hu2008instrumental,cunha2010estimating}.
Anchor-free DIF uses a global minimal-$L_1$ restriction in a MIMIC model
\citep{chen2023dif}; majority-valid IV and weighted-median MR identify scalar
causal effects~\citep{kang2016invalid,bowden2016weighted}. Our compatible-target
sets specialize partial identification~\citep{manski2003partial} to local proxy
attributions under contamination and scale uncertainty. Robust PCA separates low-rank and sparse errors
\citep{candes2011robustpca}; we instead impose a pointwise clean majority on the
attribution tensor, without identifying a latent-factor distribution or
assuming proxy independence.

\paragraph{Marginal and causal attribution.}
SHAP explains a prediction function through a chosen coalition value
\citep{lundberg2017shap}; different operationalizations can yield different
Shapley explanations~\citep{sundararajan2020many}. Marginal (often called interventional) replacement
breaks feature dependence and defines relevance relative to a background
distribution~\citep{janzing2020feature}; distribution-supported attribution
formalizes a related requirement for explaining observable classifiers
\citep{li2026distribution}. By itself, however, a reference distribution does not identify a causal
response surface from observational data. Causal Shapley instead incorporates
causal structure into the coalition value~\citep{heskes2020causal}. Our robust
aggregation result is conditional on the per-proxy value functions supplied to
it. It has a causal interpretation only when the corresponding interventional
responses are independently identified; otherwise we use the more limited term
\emph{marginal model attribution}.

\paragraph{Web preference explanations.}
LLM profile pipelines and explainers for recommendation or preference mechanisms
\citep{zhang2020explainablerec,calderon2025multidomain,langptune2026,pure2025,exp3rt2025}
produce behavioral or textual proxies or explain observed targets. We focus on a prior question that those
pipelines leave open: whether agreement among several proxy explanations is
informative about an unobserved preference, and when shared proxy bias makes
such a conclusion impossible. Recent work also shows that explanation evaluation
without ground truth is sensitive to the evaluation intervention itself
\citep{cai2026rethinking}; our real study accordingly separates a held-out
rating comparator from a controlled association-injection stress test.

%% file: sections/03_problem.tex
\section{Problem Formulation}
\label{sec:problem}

\paragraph{Latent response and reference-distribution attribution.}
Let $\pref^\ast\in\{0,1\}$ denote a latent preference that is never directly
observed, and let
$q(x)=P(\pref^\ast{=}1\mid X{=}x)$ be its latent response surface.  Let
$N=[d]=\{1,\ldots,d\}$.  For a fixed, explicit reference distribution
$P_{\rm ref}$, define for any function $h$
\begin{equation}
 v_x^h(S)=\E_{X'\sim P_{\rm ref}}
       [h(x_S,X'_{-S})],\qquad S\subseteq N.
 \label{eq:reference-value}
\end{equation}
Our local estimand and its global summary are, respectively,
\begin{equation}
 \theta_i(x)=\mathrm{Sh}_i(v_x^q),\qquad
 I_i=\E_{X\sim P_{\rm eval}}|\theta_i(X)|.
 \label{eq:local-global-target}
\end{equation}
Equation~\eqref{eq:reference-value} is the usual marginal (also called
``interventional'' in the SHAP literature) model-attribution value function.  It
does not, by itself, assign a causal meaning to $q$ or to $\theta_i(x)$.  We
treat each response as specified on the hybrid explanation domain
$\{(x_S,x'_{-S}):S\subseteq N,\ x'\in\mathrm{supp}(P_{\rm ref})\}$.
Learning it from data additionally requires support/overlap on this domain or an
explicit extrapolation model; robust aggregation does not solve off-support
identification.

\paragraph{Proxy measurement model.}
We observe proxy outcomes $Y_k$, $k\in[K]$, and can learn only their response
surfaces $m_k(x)=\E[Y_k\mid X{=}x]$. We write
\begin{equation}
 m_k(x)=a_k q(x)+b_k(x),\qquad a_k>0,
 \label{eq:proxy-measurement}
\end{equation}
where $b_k=m_k-a_kq$ is unrestricted; hence this is an accounting
decomposition, not a testable affine-in-$q$ restriction. By linearity, the
population proxy attribution $\phi_{ki}(x)=\mathrm{Sh}_i(v_x^{m_k})$ obeys
\begin{equation}
 \begin{aligned}
 z_{ki}(x)&:=\phi_{ki}(x)/a_k=\theta_i(x)+\eta_{ki}(x),\\
 \eta_{ki}(x)&=\mathrm{Sh}_i(v_x^{b_k})/a_k.
 \end{aligned}
 \label{eq:attribution-contamination}
\end{equation}
Thus our identifying restriction is imposed directly at the attribution level:
proxy $k$ is \emph{clean} for $(i,x)$ when $\eta_{ki}(x)=0$.  Independence among
the proxy outcomes is neither assumed nor used.  A feature absent from $b_k$ is
clean, although attribution-level cleanliness is the weaker condition actually
needed below.

Scale normalization in~\eqref{eq:attribution-contamination} requires $a_k$ to be
known or calibrated from external anchors; we do not identify it from the same
proxy attributions. Four regimes must be distinguished. Known positive scales
license point magnitude recovery. Estimated scales contribute normalization
error to the finite-sample bound. Externally justified scale intervals produce
compatible-target sets rather than points. With only unknown positive scales,
the raw median preserves local support and sign under a strict clean majority,
but licenses neither exact magnitudes nor cross-coordinate rankings
(Prop.~\ref{prop:a1}).

\begin{definition}[Attribution-active feature and proxy contamination]
\label{def:driver}
The global active set is $D=\{i:I_i>0\}$.  At a point $x$, proxy $k$
contaminates coordinate $i$ if $\eta_{ki}(x)\ne0$.  These definitions concern
the chosen reference-distribution attribution and make no causal claim.
\end{definition}

\paragraph{Goal.}
Given estimated proxy attributions $\widehat\phi_{ki}(x)$, we seek to recover
the \emph{local signed target} $\theta_i(x)$ first, and only then form
$I_i=\E|\theta_i(X)|$. The order matters:
\[
\operatorname{median}_k\E_X|z_{ki}(X)|
\ne \E_X|\operatorname{median}_k z_{ki}(X)|.
\]
Section~\ref{sec:theory}
characterizes exactly when the latter estimator is uniformly identified under
arbitrary attribution contamination.

%% file: sections/04_method.tex
\section{Identification Procedure and Audits}
\label{sec:method}

\paragraph{Proxy attribution equals shared signal plus contamination.}
For proxy response $m_k=a_kq+b_k$, linearity of the reference-based
value function and Shapley operator yields
$\phi_{ki}(x)=a_k\theta_i(x)+\psi_i^{b_k}(x)$, where
$\psi_i^{b_k}(x)=\mathrm{Sh}_i(v_x^{b_k})$. The decomposition is about the
population response functions; predictor and Shapley approximation errors enter
separately in Sec.~\ref{sec:theory}.

\paragraph{Aggregate locally, then summarize globally.} For every explained
user $x$, define $\hat z_{ki}(x)=\hat\phi_{ki}(x)/\hat a_k$ when scales come
from an external anchor or independent calibration data. We then aggregate
\emph{signed} calibrated proxy attributions,
\[
\begin{aligned}
\hat\theta_i^{\mathrm{med}}(x)=
\mathrm{median}_{k\in[K]}\hat z_{ki}(x),\\
\hat I_i=\frac1{n_{\mathrm{eval}}}\sum_x
|\hat\theta_i^{\mathrm{med}}(x)|.
\end{aligned}
\]
Signed aggregation preserves direction; the absolute value is used only for
global importance. The scales $a_k$ are not identified from the same
attribution tensor. Known scales yield point magnitudes; independently estimated
scales add the normalization term in Eq.~\eqref{eq:total-error}; and externally
justified scale intervals yield the exact pointwise set in
Prop.~\ref{prop:a1}. If no calibration is available, we median-aggregate raw
$\hat\phi_k$ and report support, direction, and an uncalibrated consensus
score---not latent magnitude or feature ranking. A scale shared across points
is outside our pointwise analysis under interval-scale uncertainty.

\paragraph{Which output to report.}
Evidence determines the output. If negligible contamination is externally
justified and efficiency is primary, the mean is an efficiency-oriented
summary. If arbitrary contamination by a strict minority is plausible and a
strict clean majority is externally justified, the median yields point support
and direction; magnitude additionally requires anchored positive scales. With
only an $(r,\epsilon)$ budget or scale intervals, report the exact compatible
set from Props.~\ref{prop:sensitivity} and~\ref{prop:a1}: excluding zero licenses
a sign, containing zero abstains, and emptiness flags the joint declaration.
Without external support, report consensus and sensitivity; agreement alone
cannot choose among these branches.

Under coordinatewise arbitrary contamination, the median needs no joint proxy
model and tolerates any strict minority of arbitrarily contaminated proxies;
Thm.~\ref{thm:identif} proves the corresponding finite-$K$ identification
boundary. Low-rank--sparse and label-first alternatives impose different
structures or objectives; the controlled benchmark compares them on common data.

\paragraph{Reference-distribution value estimation.} We fit $f_k$ to the
conditional proxy response and explain it with a self-implemented Kernel SHAP.
Off-features are drawn from a declared marginal reference distribution $P_{\rm ref}$;
we therefore call the result \emph{marginal} or \emph{model-interventional}
attribution, not an identified causal effect. Ordinarily, coalitions are exact
when $2^d\le4096$ and Shapley-kernel--sampled otherwise; the numerical-error
audit deliberately samples at $d=8$ to isolate coalition Monte Carlo error.
Efficiency-constrained weighted least squares is solved in closed form via its
KKT system (machine-precision efficiency).

The executable sequence is explicit: fit each fixed proxy response, compute
local attributions under the declared $P_{\rm ref}$, normalize only with
external scales, take the signed coordinatewise median, and only then form
global summaries.

\paragraph{Audit interpretation.} The strict-majority condition is structural:
proxy agreement does not verify it. Cross-modal design diversifies measurement
pathways, but external anchors, validation data, or a justified channel model
must support the cleanliness and scale budgets. Otherwise, outputs are
observable consensus and sensitivity summaries.

%% file: sections/05_theory.tex
\section{Theoretical Analysis}
\label{sec:theory}

We first expose proxy leakage, then give a sharp attribution-level
identifiability boundary. The result concerns the local signed estimand
$\theta_i(x)$; global importance $I_i$ follows only after local recovery.
The statements, constructions, and proof sketches below expose the decisive
mechanisms.

Building on Shapley linearity, median trapping, and concentration, we provide a
response-valid indistinguishability construction, exact compatible-target sets
for known and interval scales, and an attribution-error interface that
propagates uncertainty to local, global, and support recovery.

\begin{theorem}[Single-proxy leakage]
\label{thm:leak}
Under~\eqref{eq:proxy-measurement}, the population attribution of proxy $k$ is
\begin{equation}
 \phi_{ki}(x)=a_k\theta_i(x)+\psi^{b_k}_i(x),
 \qquad \psi^{b_k}_i(x)=\mathrm{Sh}_i(v_x^{b_k}).
 \label{eq:leakage}
\end{equation}
If $b_k$ is invariant to its $i$th argument throughout the hybrid explanation
domain, then $\psi^{b_k}_i(x)=0$; feature-dependent bias may induce nonzero
leakage.  With unrestricted $b_k$, one proxy cannot uniformly identify
$\theta_i(x)$.
\end{theorem}

\noindent\emph{Proof sketch.}
Both the reference-distribution value function and the Shapley operator are
linear, giving~\eqref{eq:leakage}.  If $b_k$ is invariant to its $i$th argument,
every coalitional marginal contribution of $i$ is zero. Conversely, choose an
interior response and transfer a sufficiently small bounded component between
$q$ and $b_k/a_k$; $m_k$ is unchanged.\hfill$\square$

\begin{corollary}[Non-differential bias]
\label{cor:nondiff}
If $b_k(x)$ is constant in $x$, its attribution vanishes.  A calibrated proxy
then identifies $\theta(x)$; without calibration it identifies it only up to the
positive factor $a_k$.
\end{corollary}

For ordered scalars $u_{(1)}\le\cdots\le u_{(K)}$, we use the signed median
\begin{equation}
 \operatorname{med}(u)=
 \begin{cases}
 u_{((K+1)/2)},&K\text{ odd},\\
 (u_{(K/2)}+u_{(K/2+1)})/2,&K\text{ even}.
 \end{cases}
 \label{eq:median-convention}
\end{equation}

\begin{theorem}[Sharp strict-majority identifiability]
\label{thm:identif}
Fix an integer contamination budget $0\le r\le K$.  Suppose that, for every $(i,x)$,
$z_{ki}(x)=\theta_i(x)$ for at least $K-r$ proxies, while the remaining values
are arbitrary.  If $2r<K$, then
\begin{equation}
 \theta_i(x)=\operatorname{med}_{k\in[K]}z_{ki}(x),\qquad
 I_i=\E_X\left|\operatorname{med}_k z_{ki}(X)\right|.
 \label{eq:median-recovery}
\end{equation}
Conversely, if $2r\ge K$, two distinct latent attributions can generate the same
$K$ proxy attributions while each contaminates at most $r$ proxies.  Hence no
estimator uniformly identifies $\theta_i(x)$ under arbitrary contamination at
or beyond this boundary.
\end{theorem}

\noindent\emph{Proof sketch.}
When $2r<K$, strictly more than half of the entries equal $\theta_i(x)$, so the
central order statistic---both central statistics when $K$ is even---equals
$\theta_i(x)$.  For $2r\ge K$, take disjoint sets $A,B$ of size $K-r$ and an
observed vector equal to $t_0$ on $A$ and $t_1\ne t_0$ on $B$.  It is compatible
with either latent value after changing at most $r$ entries. This ambiguity is
response-valid: for one feature $u\in[-1,1]$, a reference with
$\E_{\rm ref}U=0$, evaluation
$u=1$, and $|z_k|,|t_j|<1/2$, set $q_j(u)=1/2+t_ju$ and
$b_{k,j}(u)=(z_k-t_j)u$. Both candidates yield the same bounded proxy response
$m_k(u)=1/2+z_ku\in[0,1]$ but latent attribution $t_j$. For this lower-bound
construction, give $U$ the same distribution in both worlds and choose the same
conditional coupling of the proxies; for example, conditional on $U=u$, draw
$Y_1,\ldots,Y_K$ independently with
$Y_k\sim\operatorname{Bernoulli}(m_k(u))$. Since every $m_k$ is identical
between the worlds, the complete observed laws of
$(U,Y_1,\ldots,Y_K)$ coincide. Appending arbitrarily correlated nuisance
features ignored by all responses embeds the construction in any $d>1$;
$P_{\rm ref}$ need not factorize.\hfill$\square$

The clean set may vary by user and coordinate, without proxy independence or
disjoint bias supports. For $K=2$, the boundary permits only $r=0$.

\begin{proposition}[Sharp approximate risk and compatible-target set]
\label{prop:sensitivity}
Let $r\in\{0,\ldots,K\}$ and $\epsilon\ge0$. Over pairs $(z,t)$ for which at
least $K-r$ entries of $z$ are within $\epsilon$ of $t$, the unrestricted-target
minimax risk is
\begin{equation}
 R^\star_{K,r,\epsilon}=\inf_T\sup_{(z,t)}|T(z)-t|
 =\begin{cases}\epsilon,&2r<K,\\ \infty,&2r\ge K.\end{cases}
 \label{eq:minimax-risk}
\end{equation}
where the infimum is over $T:\mathbb R^K\to\mathbb R$. The median attains the
first value. On a known finite interval $[L,U]$, the
risk is $\min\{\epsilon,(U-L)/2\}$ below the boundary and $(U-L)/2$ at or
beyond it, attained by projecting the median or using the midpoint. Thus
binary-response attributions in $[-1,1]$ have finite risk even without point
identification.

For observed normalized attributions $z=(z_1,\ldots,z_K)$ at a fixed $(i,x)$
and a declared target domain $\mathcal T\subseteq\mathbb R$, define
\begin{equation}
 \mathcal C^{\mathcal T}_{r,\epsilon}(z)=
 \{t\in\mathcal T:\,|\{k:|z_k-t|\le\epsilon\}|\ge K-r\}.
 \label{eq:sensitivity-set}
\end{equation}
This exact set uses $\mathcal T=[-1,1]$ for binary responses (and $\mathbb R$
when the superscript is omitted). A nonempty bounded set has conditional minimax
radius half its diameter. If $2r<K$, its diameter is at most $2\epsilon$ and the
median is within $\epsilon$ of every member; at or beyond the boundary its
unrestricted diameter can be arbitrarily large over admissible $z$ (and is
unbounded for $r=K$). It is a sensitivity report for stated
$(r,\epsilon)$, not an estimate of them from agreement.
\end{proposition}

\begin{corollary}[Externally validated probabilistic cleanliness]
\label{cor:audit}
Index $N_{\rm pair}$ prespecified user--coordinate pairs by $n$. On a common
proxy-channel generation space let
$C_{nk}=\mathbf 1\{|z_{nk}-\theta_n|\le\epsilon\}$. Suppose
$\{C_{nk}\}_{k=1}^K$ are independent for each fixed $n$ and
$\inf_{n,k}\Pr(C_{nk}=1)\ge p_0:=1/2+\gamma$, where
$0<\gamma\le1/2$. Define
$\beta_K(\gamma)=\Pr\{\operatorname{Bin}(K,p_0)\le\lfloor K/2\rfloor\}$.
Then, with probability at least
$[1-N_{\rm pair}\beta_K(\gamma)]_+$, every pair has a strict approximate-clean
majority and its median error is at most $\epsilon$ (zero when $\epsilon=0$).
Here $[u]_+=\max\{u,0\}$, $\beta_K$ is the exact worst-case per-pair failure
probability over the allowed independent channels, and
$\beta_K(\gamma)\le\exp(-2K\gamma^2)$. No independence across pairs is needed.
The probability premise must come from independent latent-target validation, a
known randomized measurement mechanism, or a proved channel model---never from
agreement among the analyzed proxies. If, conditional on an external validation
event $V$ with $\Pr(V)\ge1-\delta_{\rm val}$, the channels retain within-pair
independence and
$\inf_{n,k}\Pr(C_{nk}=1\mid V)\ge p_0$, the unconditional lower bound is
$[1-\delta_{\rm val}-N_{\rm pair}\beta_K(\gamma)]_+$.
\end{corollary}

\paragraph{Conditional falsification and abstention.}
For estimated inputs, use
$\mathcal C^{\mathcal T}_{r,\epsilon+\varepsilon_{\rm attr}}(\widehat z)$.
If one set of at least $K-r$ proxies satisfies both declared structural and
estimation errors, the triangle inequality places its target in this set.
Emptiness therefore falsifies that joint declaration; a nonempty set identifies
a sign only if all members share one strict sign, and never certifies the
budgets themselves.

\begin{proposition}[$K=2$ under support separation]
\label{prop:k2}
For $P_{\rm eval}$-almost every $x$, let $D_x=\{i:\theta_i(x)\ne0\}$ and
$B_k(x)=\{i:\eta_{ki}(x)\ne0\}$. If $B_1(x)\cap B_2(x)=\varnothing$ and
$D_x\cap(B_1(x)\cup B_2(x))=\varnothing$, then, for every coordinate,
\begin{equation}
 \min_{k\in\{1,2\}}|z_{ki}(x)|=|\theta_i(x)|.
 \label{eq:k2-min}
\end{equation}
Thus $I_i=\E_X\min_k|z_{ki}(X)|$ under this stronger separation model, not a
$50\%$ robustness guarantee.
\end{proposition}

\begin{theorem}[Uniform error propagation]
\label{thm:finite}
Let $X_1,\ldots,X_n\stackrel{\rm iid}{\sim}P_{\rm eval}$ and
$\widehat z_{ki}(x)=\widehat\phi_{ki}(x)/\widehat a_k$. Suppose that for some
$r\in\{0,\ldots,K\}$ with $2r<K$, the strict-majority condition of
Thm.~\ref{thm:identif} holds for every coordinate and
$P_{\rm eval}$-almost every $x$. Suppose further that, with probability at least
$1-\delta_{\rm attr}$, simultaneously for all $i\in[d]$, $j\in[n]$, and every
proxy $k$ that is population-clean at $(i,X_j)$,
\begin{equation}
 \begin{aligned}
 |\widehat z_{ki}(X_j)-z_{ki}(X_j)|
 &\le\underbrace{\varepsilon_{\rm model}}_{\text{response fit}}
     +\underbrace{\varepsilon_{\rm bg}}_{\text{reference MC}}\\
 &\quad+\underbrace{\varepsilon_{\rm coal}}_{\text{coalition MC}}
     +\underbrace{\varepsilon_{\rm scale}}_{\text{normalization}}
 =:\varepsilon_{\rm attr}.
 \end{aligned}
 \label{eq:total-error}
\end{equation}
The remaining proxy estimates may be arbitrary. Then, simultaneously for all
$i,j$,
\[
|\widehat\theta_i(X_j)-\theta_i(X_j)|
\le\varepsilon_{\rm attr},
\quad
\widehat\theta_i(X_j)=\operatorname{med}_k\widehat z_{ki}(X_j).
\]
If $|\theta_i(X)|\le M$ almost surely and
\[
\widehat I_i=\frac{1}{n}\sum_{j=1}^n
|\widehat\theta_i(X_j)|,
\]
then, with probability at least
$1-\delta_{\rm attr}-\delta_{\rm eval}$,
\begin{equation}
 \max_i|\widehat I_i-I_i|\le
 \tau:=\varepsilon_{\rm attr}+
 M\sqrt{\frac{\log(2d/\delta_{\rm eval})}{2n}}.
 \label{eq:global-error}
\end{equation}
With the convention $\min_{i\in\varnothing}I_i=+\infty$, let
$\Delta=\min_{i\in D}I_i$. Consequently, if $\Delta>2\tau$, the strict-threshold
estimate $\widehat D=\{i:\widehat I_i>\tau\}$ recovers $D$ exactly, including
when $D=\varnothing$.
\end{theorem}

The four terms separately audit response fitting, reference and coalition
Monte Carlo, and scale calibration; $r$ and $\epsilon$ require external
validation, not proxy agreement.

\begin{proposition}[Interval-calibrated and unknown positive scales]
\label{prop:a1}
Let $K\ge1$, $r\in\{0,\ldots,K\}$, $\epsilon\ge0$, and
$\mathcal T\subseteq\mathbb R$. Fix $(i,x)$ and use only its raw attributions
$p_k=\phi_{ki}(x)$. Let external calibration give
$A_k=[\underline a_k,\overline a_k]\subset(0,\infty)$ and set
\begin{align}
 J_k(p_k)&=\big[\min_{a\in A_k}p_k/a-\epsilon,
                     \max_{a\in A_k}p_k/a+\epsilon\big],\nonumber\\
 \mathcal C^{\mathcal T}_{r,\epsilon}(p;A)
 &=\{t\in\mathcal T:\sum_k\mathbf 1\{t\in J_k(p_k)\}\ge K-r\}.
 \label{eq:scale-compatible-set}
\end{align}
This is exactly the pointwise target set compatible with some $a_k\in A_k$, at
most $r$ arbitrary proxies, and normalized clean error at most $\epsilon$. Let
$w_k=|p_k|(\underline a_k^{-1}-\overline a_k^{-1})+2\epsilon$ and write their
order statistics as $w_{(1)}\le\cdots\le w_{(K)}$. Whenever the compatible-target set
is nonempty and $2r<K$,
\begin{equation}
 \operatorname{diam}\mathcal C^{\mathcal T}_{r,\epsilon}(p;A)
 \le \min\{\operatorname{diam}(\mathcal T),w_{(2r+1)}\}.
 \label{eq:scale-compatible-diameter}
\end{equation}
Singleton scales recover Eq.~\eqref{eq:sensitivity-set}. The set is pointwise;
one scale vector shared across points would define a tighter joint feasibility
problem, which we do not analyze. Scale intervals require external evidence. If random intervals cover all true scales
simultaneously with probability at least $1-\delta_{\rm scale}$, conditional
claims inherit that failure term; it is not subtracted again when already
included in $\delta_{\rm attr}$.

If $0<a_{\min}\le a_k\le a_{\max}$ and a strict majority of raw attributions
satisfy $\phi_{ki}=a_k\theta_i$, their median has the support and sign of
$\theta_i$ and obeys
\begin{equation}
 a_{\min}|\theta_i(x)|\le
 |\operatorname{med}_k\phi_{ki}(x)|\le
 a_{\max}|\theta_i(x)|.
\end{equation}
Thus the raw coordinatewise median alone guarantees neither exact magnitudes
nor cross-coordinate rankings without calibration.
\end{proposition}

\paragraph{Proof roadmap.}
For Prop.~\ref{prop:sensitivity}, a common observation compatible with two
targets gives the minimax lower bounds; below the boundary, two supporting sets
of size $K-r$ intersect, giving diameter $2\epsilon$, while the median attains
the upper bound. Cor.~\ref{cor:audit} couples each clean count to
$\operatorname{Bin}(K,1/2+\gamma)$, uses its exact lower tail, and unions over
prespecified pairs; Hoeffding gives the exponential relaxation.
Prop.~\ref{prop:k2} splits on $i\in D_x$:
both leakages vanish there, whereas off $D_x$ disjoint bias supports leave one
proxy at zero. For Thm.~\ref{thm:finite}, a clean majority traps the median in
the $\varepsilon_{\rm attr}$ interval; the reverse triangle inequality and a
Hoeffding--union bound over coordinates yield $\tau$. Finally, two targets in
the scale-uncertain compatible-target set share at least $K-2r$ supporting intervals, implying
$|t-t'|\le w_{(2r+1)}$ in Prop.~\ref{prop:a1}.

\begin{corollary}[Transfer to externally identified coalition values]
\label{cor:causal}
If an external causal design identifies all latent and proxy coalition values
with $v_{k,x}(S)=a_kv_x^\star(S)+v_x^{b_k}(S)$, our results transfer to their
Shapley values, subject to their respective cleanliness, scale, and error
assumptions. A full-intervention surface alone need not identify every
coalition value; fitting $\E[Y_k\mid X]$ supports only our model-attribution
interpretation.
\end{corollary}

\paragraph{Scope.}
Causality-aware Shapley work~\citep{heskes2020causal,janzing2020feature} studies
observable targets under causal assumptions.  Here the distinct question is
robust identification of a latent target's \emph{attribution vector} from
proxy-attribution contamination.  The signed median attains the sharp
$2r<K$ point-identification boundary; the result does not identify the latent
variable distribution itself and does not turn observational proxy models into
causal models.

%% file: sections/06_experiments.tex
\section{Experiments}
\label{sec:experiments}

We use controlled and real-data evidence. A known-truth benchmark maps the null,
boundary, shared-majority, and bias-strength regimes across eleven methods. A
direct attribution-tensor audit verifies the compatible-set interface: at the
correct $\epsilon=.01$ budget it attains $100\%$ truth coverage,
$95.40\%/97.05\%$ active-driver sign decisions, and zero false signs. A
temporally ordered Amazon study asks: \textbf{RQ1} later-response alignment;
\textbf{RQ2} median versus behavior; \textbf{RQ3} compatible-set measurement
readiness; and \textbf{RQ4} a controlled perturbation on real covariates. Its
anchors are observable responses, never latent preference ground truth.

\subsection{Controlled Known-Truth Benchmark}
\label{sec:synthetic-benchmark}

\paragraph{Design and evidential status.}
We generate independent train and evaluation covariates in eight dimensions
with two latent drivers and a separate bias coordinate. An additive-probability
DGP and a nonlinear logistic-link DGP have known latent responses, scales, clean
identities, and attributions. With $X\sim{\rm Unif}([-1,1]^8)$ and contaminated
indicator $c_k$, they use
$q_{\rm add}=.5+.12x_1-.10x_2$,
$p_k=q_{\rm add}+.18c_k\lambda x_3$, and
$p_k=\sigma(1.2x_1-x_2+1.5c_k\lambda x_3)$ for the logistic stress; all scales
equal one. Each cell fits on 6,000 cases and explains 200 independent cases
exactly. For $K\in\{3,4,5,7\}$ we scan every contamination
count $m=0,\ldots,K$ and strength $\lambda\in\{0.5,1,1.5\}$, comparing eleven
fixed estimator implementations over the same ten seeds; the clean-proxy oracle is
undefined at $m=K$. This full grid is exploratory and
descriptive: its seed intervals show variation, not a familywise confirmatory
claim. After inspecting this exploratory grid, but before any of twenty unseen
confirmation seeds was run, we fixed the paired median--RPCA contrast at
$K=5,m=2,\lambda=1$, the most contaminated $K=5$ cell inside the
strict-majority guarantee. The complete design contains 138 scenarios and
1,494 method--scenario summaries; only the separately frozen two-DGP contrast
is treated as confirmatory.

\begin{figure*}[t]
\centering
\includegraphics[width=0.92\textwidth]{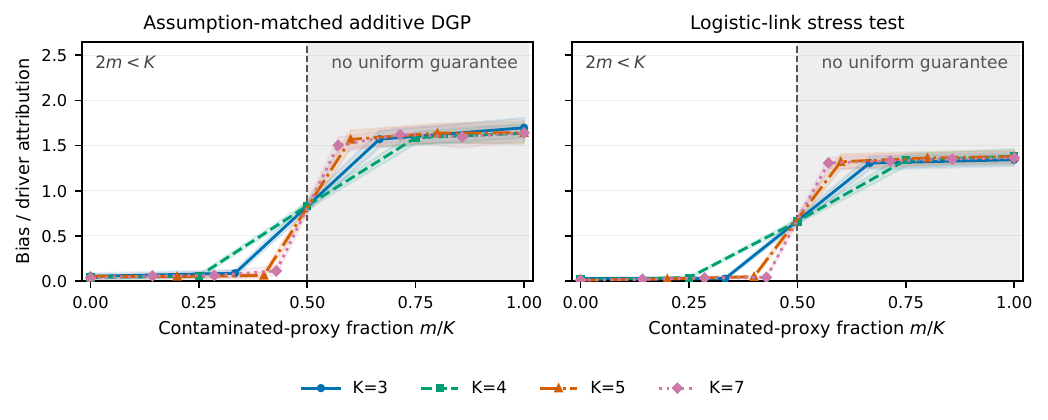}
\caption{\textbf{Controlled strict-majority phase diagram} at $\lambda=1$.
The vertical axis reports bias leakage: mean absolute attribution on the
designated bias coordinate divided by the average mean absolute attribution
over the two true driver coordinates. It is shown for the signed local median
at every contamination count and $K\in\{3,4,5,7\}$ in the additive and
logistic-link DGPs. Bands are
pointwise 95\% Student-$t$ intervals over ten exploratory seeds. Leakage remains
small when $2m<K$ and rises under the shared-direction construction when the
uniform guarantee ends. The theorem is worst-case: losing the guarantee does
not require every beyond-boundary instance to fail.}
\Description{Two panels plot median bias leakage against contamination count
for four proxy counts. Leakage is low under a strict clean majority and rises
at the first boundary-violating count.}
\label{fig:synthetic-phase}
\end{figure*}

\paragraph{Boundary, null, and strength behavior.}
Figure~\ref{fig:synthetic-phase} supplies both favorable and adverse regimes.
At $m=0$, the mean can be more efficient than the median because there is no
contamination to reject: at $K=3,\lambda=1$, mean/median normalized local
$\ell_1$ error was $.179/.209$ (additive) and $.085/.102$ (logistic). At the
below-boundary $m=1$ cell, this ordering reversed to $.426/.247$ and
$.301/.100$. At the first $m$ satisfying $2m\ge K$, leakage increases sharply;
$K=3,m=2$ is an explicit shared-majority negative control rather than
independent canceling noise. For $K=5$, median leakage across
$\lambda=(.5,1,1.5)$ was $(.082,.063,.097)$ and $(.051,.049,.048)$ at $m=2$
(additive/logistic), versus $(.741,1.569,2.379)$ and $(.648,1.322,1.931)$ at
$m=3$. Thus the strength sweep preserves the qualitative boundary while
exposing DGP- and strength-dependent error magnitude.

\paragraph{Eleven-method comparison and confirmation.}
The alternatives deliberately encode different structures: robust locations,
rank-one and low-rank--sparse reconstructions, a label-first response, and a
nondeployable clean-proxy oracle. They are comparable here because latent truth
is known, not because they are interchangeable objectives on Amazon. On the
twenty unseen confirmation seeds, mean bias leakage of the local median was
$.115$ in the additive DGP and $.050$ in the logistic DGP, versus $.142$ and
$.058$ for RPCA. The frozen confirmation-cohort paired median-minus-RPCA effects were $-.0277$
and $-.0079$. Treating the two fixed DGP contrasts as one reporting family,
their two-sided Bonferroni family-two 95\% intervals were
$[-.0379,-.0175]$ and $[-.0116,-.0042]$; both remain below zero. Both phases
used the same implementations, per-seed sample sizes, focal-cell definition,
and leakage metric; only their disjoint seed cohorts differed. The additive
median intervals $[.030,.096]$ (discovery) and $[.083,.147]$ (confirmation)
overlap slightly, and the median--RPCA effect replicated in direction
($-.0188$ versus $-.0277$). Thus the $.063$--$.115$ absolute-level shift is
between-cohort seed variation; the confirmation claim remains the frozen
paired contrast. These confirmation statements apply only to the fixed
$K=5,m=2,\lambda=1$ cell;
they do not convert the exploratory phase grid into a dominance claim.

\begin{table}[t]
\centering
\small
\resizebox{\columnwidth}{!}{%
\input{\WSDMRevisionRoot/main_synthetic_confirmation.tex}}
\caption{Controlled confirmation-cohort bias leakage and seed-paired
median-minus-RPCA effects at $K=5,m=2,\lambda=1$ (20 unseen seeds; lower is
better). Family-two intervals apply a reporting-stage Bonferroni correction
across the two fixed DGP contrasts. This is the only frozen confirmatory
method contrast; the remaining methods belong to the descriptive grid.}
\label{tab:synthetic-confirmation}
\end{table}

\paragraph{Direct attribution-tensor audit.}
To isolate the attribution-specific mechanism from response fitting, a
separate known-truth suite constructs the signed local attribution tensor
directly ($K=5$, true $r=2$; ten seeds, 512 cases and eight coordinates per
seed). Across all ten nonzero $\epsilon\times b$ cells under the correct budget,
truth coverage was $100\%$, no set was empty, every diameter bound held, and
mean diameter divided by $\epsilon$ was $.997$--$1.351$. Table~\ref{tab:synthetic-compatible}
shows the $\epsilon=.01$ slice and deliberate post-primary misspecification
checks. Halving the declared tolerance reduced coverage, while understating
$r$ made every set empty in this constructed stress regime. Emptiness
falsifies only the joint budget and structure declaration; it does not locate
the failed component, and nonemptiness does not validate the declaration.

\begin{table}[t]
\centering
\scriptsize
\resizebox{\columnwidth}{!}{%
\input{\WSDMRevisionRoot/main_synthetic_compatible_sets.tex}}
\caption{Direct known-truth compatible-target-set audit at true $(r,\epsilon)=(2,.01)$.
$b$ is the shared clean-proxy offset as a fraction of $\epsilon$. Coverage,
diameter, and emptiness use all $512\times8$ case--coordinate cells per seed;
decision/false uses the two active drivers. Ten seeds are the replicates.
Misspecified rows are descriptive post-primary stress tests; ``--'' means all
sets are empty.}
\label{tab:synthetic-compatible}
\end{table}

At the strongest positive-scale stress, the raw median retained exact support
and direction but had magnitude ratio $1.25$ and global relative $\ell_1$ error
$.250$. Known-scale normalization recovered zero error; normalization from 256
independent anchors reduced global error to $.00384$ (seed-level 95\% CI
$[.00152,.00617]$). This audit isolates attribution-tensor behavior and does
not establish response-function realizability or Amazon budget validity.

\subsection{Amazon Study: Pilot-Supported, Formal-Test-Blind Design}

\paragraph{Data and temporal separation.}
We use the frozen 100,000-user \texttt{CDs\_and\_Vinyl} subset of Amazon Reviews
2023~\citep{amazonreviews2023} and its item metadata. For every eligible user,
the first 20 chronologically ordered interactions are divided into disjoint
history, proxy, and future windows of 12, 4, and 4 events. The formal cohort
contains 5,000 users, split by fixed hash ranking into 3,000 response-training,
500 scale-fitting, 500 error-calibration, and 1,000 test users. It is disjoint
from the pilot, engineering, and other study cohorts. Pop, Rock, Jazz, and
Country were selected by a support-only pilot rule and then frozen. The
ordering is temporal within each user; the hash-partitioned roles need not
occupy nonoverlapping global calendar periods. Thus the design evaluates
forward response within users rather than ecosystem-wide calendar drift.

The seven covariates are history-only activity, rating, diversity, timing,
entropy, and trend summaries. The behavior proxy records target engagement in
the proxy window. Qwen2.5-7B-Instruct-4bit and
Meta-Llama-3-8B-Instruct-4bit profiles are generated offline from the same
history packet using greedy decoding (temperature 0, top-$p$ 1, at most 200
tokens) and parsed by a fixed schema. The primary comparator,
\texttt{future\_rated}, records a target-mapped rating of at least four stars in
the future window; proxy-window rating is secondary. Neither features nor LLM
prompts use future-window information. A successfully parsed empty profile is
retained and maps deterministically to zero for every target-specific binary
response; it is neither dropped nor marked unavailable.

All 5,000 records per model completed, while exact-filter repairs were requested
for $41.6\%/36.7\%$ of Qwen/Llama generations and $24.1\%/23.4\%$ parsed as
empty. Because both LLM views use the same history packet, common contamination
is a relevant stress case: if both views were contaminated, $K=3,r=2$ would lie
beyond the strict-majority guarantee. We therefore treat Amazon as a
conservative observable-surface audit, use the controlled $K=3,m=2$ condition
to characterize the corresponding failure mode, and do not invoke
Cor.~\ref{cor:audit}, whose externally validated within-pair independence and
cleanliness premises are not established here.

\paragraph{Models, attributions, and method inventory.}
For every supported target--response pair, we fit the same deterministic,
unweighted LightGBM classifier (160 trees, learning rate $.05$, 31 leaves,
seed 42) on the training role. Local attributions use all $2^7=128$ coalitions
per user--response pair and a fixed 128-user training background. Exact
enumeration is exponential in $d$, motivating sampled coalitions at larger
dimension. Aggregation is coordinate-wise on signed local attributions before
global absolute importance.
The unscaled three-view mean and median use behavior, Qwen, and Llama. Positive
point scales, when available, are fitted only on the scale role against the
future-rated response as origin-constrained slopes
$\widehat a_k=\sum_{i,j}\phi^F_{ij}\phi^k_{ij}/
\sum_{i,j}(\phi^F_{ij})^2$; a separate role supplies the compatible-target-set error radius
and conditional-sign margin. The eleven prespecified Amazon entries comprise
ten point-attribution methods and one compatible-set output; their scale gates
differ, and no all-method dominance claim is made.

On the 500-user error-calibration role, the radius is the 476th ordered
per-user maximum absolute discrepancy across all three calibrated proxies and
seven coordinates. The target-specific CSE margin is the $.25$ quantile of
strictly positive absolute comparator attributions on that same role.

\paragraph{Test protocol.}
All cohorts, targets, features, learners, metrics, bootstrap seeds, and primary
analysis outputs were fixed before test access. The primary paired contrast is the
unscaled K=3 median minus its behavior-only constituent: an incremental
multi-view comparison, not two independent evidence streams.
Direction-normalized attribution error (DNAE) uses
all 1,000 test users; conditional sign error (CSE) uses a method-independent
effective set fixed by the comparator. Lower is better. Each target--metric cell
uses 10,000 user-cluster bootstrap resamples, with ordinary and fixed-family
Bonferroni intervals. The test role was evaluated once. Frozen aggregate assets
and the explicitly labeled complete reporting extension are generated without
manual transcription or within-table cell selection.

Precisely, for a seven-vector $v$, let $u(v)=v/\lVert v\rVert_1$ when
$\lVert v\rVert_1>10^{-12}$ and zero otherwise. For method $a$, comparator
$F$, and user $i$, DNAE is
$7^{-1}\sum_{j=1}^7|u(\phi_i^a)_j-u(\phi_i^F)_j|$. CSE is the proportion of
sign disagreements with $F$ among coordinates satisfying
$|\phi^F_{ij}|\ge\delta_c$; its user denominator is the fixed effective set
containing at least one eligible coordinate.

\subsection{RQ1--RQ2: Fixed Future-Response Alignment}

Table~\ref{tab:rq1-rq2-primary} reports every primary cell. The effect is median
minus behavior, so a negative value favors the median. Wins, ties, and losses
use user-level paired differences; an empty fixed CSE effective set produces an
unavailable cell rather than a changed denominator.

\begin{table*}[t]
\centering
\small
{\scriptsize\textbf{(a) Frozen primary paired comparison}}\par
\resizebox{\textwidth}{!}{%
\input{\WSDMRevisionRoot/main_rq1_rq2_absolute_effects.tex}}
\vspace{2pt}
{\scriptsize\textbf{(b) Complete point-method inventory (descriptive)}}\par
{\scriptsize
\resizebox{\textwidth}{!}{%
\input{\WSDMRevisionRoot/appendix_amazon_point_method_inventory.tex}}}
\caption{Amazon point-attribution results (lower is better). Panel (a) is the
frozen paired $K=3$ median-versus-behavior comparison; Bonferroni intervals
cover its eight cells. Panel (b) lists every frozen point method without cell
filtering or new inference. Rows differ in scale or objective, $K=4$ is
sensitivity only, and the primary contrast remains panel (a).}
\label{tab:rq1-rq2-primary}
\end{table*}

All eight point estimates favored the median, with heterogeneous uncertainty.
On the absolute scale, point-estimate reductions were $6.5\%$--$13.0\%$ for CSE
and $0.7\%$--$7.6\%$ for DNAE.
Bonferroni intervals excluded zero for six contrasts: CSE decreased for every
target by $.021$--$.054$, while DNAE decreased for Pop ($-.0106$,
$[-.0138,-.0074]$) and Country ($-.0041$, $[-.0074,-.0006]$). Rock and Jazz
DNAE intervals crossed zero. Panel~(b) of
Table~\ref{tab:rq1-rq2-primary} is a
post-test, reporting-only complete inventory giving point estimates for all ten
frozen point methods. Relative to the best of the four single views, the K=3
median had lower loss in four of eight target--metric cells; the descriptive
K=4 median including rating did so in seven, with Rock CSE as the exception.
The unscaled K=3 mean was also below the K=3 median in seven of eight point
cells. This post-test descriptive pattern matches the controlled
efficiency--robustness tradeoff: the mean was more efficient at $m=0$, whereas the ordering
reversed at $m=1$. It neither establishes mean dominance nor diagnoses
contamination; the reporting rule in Sec.~\ref{sec:method} distinguishes an
efficiency-oriented mean from the median's worst-case guarantee.
The latter is the ordinary even-sample median (the mean of the two central
values), has no interval or new inferential claim, and does not overcome the
nonidentifiability of two jointly contaminated LLM views.

Held-out AUC and Brier skill were $.597$ and $-1.20\%$ for Pop, and $.536$ and
$-5.26\%$ for Rock. The corresponding pairs were $.542$ and $-11.62\%$ for Jazz,
and $.634$ and $-4.39\%$ for Country. Negative skill means that Brier loss exceeded the
fixed train-prevalence baseline by $.0030$--$.0127$, not that a surface is
random. This calibration audit bounds the interpretation: Amazon establishes
alignment to weak fitted future-response surfaces, not predictive superiority,
comparator validation, latent-preference recovery, or latent-scale recovery.

\subsection{RQ3: Compatible-Target-Set Measurement-Readiness Audit}

RQ3 asks whether current measurements, calibrations, and budgets license stable
comparator-relative signs. A compatible set signs only when all values agree;
containing zero abstains, and emptiness flags the joint scale, error, and
contamination specification. This is a budget-conditional readiness audit, not
another point estimator.

\begin{table*}[t]
\centering
\scriptsize
\resizebox{\textwidth}{!}{%
\input{\WSDMRevisionRoot/main_rq3_compatible_sets_exact.tex}}
\caption{RQ3 comparator-relative point-scale compatible-target sets. Coverage
is an all-coordinate user event with an exact Clopper--Pearson 95\% interval;
sign and abstention use all coordinate sets. Discordance uses eligible decisions
under the fixed comparator margin; the comparator is not latent truth, and
``--'' means no eligible decision.}
\label{tab:rq3-compatible}
\end{table*}

At $r=0$, simultaneous comparator coverage was $.935$--$.961$, strict signs
were licensed for $.53\%$--$1.41\%$ of coordinates, and abstention was
$98.54\%$--$99.47\%$. At $r=1$, coverage rose to $.990$--$1.000$, strict-sign
rates were at most $.16\%$, and abstention was $99.84\%$--$100\%$. The readiness
verdict is clear: under the current calibrated radii and declared budgets, this
three-proxy measurement system rarely supports comparator-relative sign claims.
Abstention prevents unsupported point conclusions; it is not evidence of zero
preference.

Comparator discordance used 34--95 eligible decisions per target at $r=0$;
Rock and Country were $22/34$ and $24/46$, and $r=1$ was undefined for three
targets. These small-denominator comparisons diagnose the rare licensed
decisions against fitted comparators; they are not latent-direction error rates.

Calibrated radii were $.651$ (Pop), $1.756$ (Rock), $.345$ (Jazz), and $.697$
(Country); CSE margins were $.009$, $.010$, $.003$, and $.006$. In the same
order, mean set diameters at $r=0$ were $1.142$, $1.957$, $.600$, and $1.236$;
at $r=1$ they were $1.302$, $1.999$, $.690$, and $1.394$.

A descriptive calibration-role decomposition localizes Rock's $1.756$ radius
to Llama: its comparator-anchored positive scale was $.0938$, so normalization
amplified its discrepancies; that view supplied the per-user maximum for
$453/500$ calibration users and the 476th-order cutoff. The resulting
$1.957/1.999$ mean diameters at $r=0/1$ rule out deploying Rock's rare
strict-sign outputs. Any actionable sign still
requires external support for target semantics, scales, and $(r,\epsilon)$; here
the first remedy is a stronger independent scale anchor for Llama, or its
replacement within a clean-majority-capable proxy set.

\subsection{RQ4: Single-Regime Real-Covariate Mechanism Check}

For each target, two proxy labels copy the unperturbed behavior response and
acquire distinct feature-conditioned associations. From a disjoint 128-user
training subset, we choose two behavior-anchor drivers and two weak injection
coordinates; above each injection coordinate's $.70$ quantile, an unperturbed
zero label flips with probability $.60$. All models are fitted before test
access. ``Unperturbed'' means unaltered by this construction, not latent truth.
We compare the coordinatewise mean and median of the three attribution vectors.
This is one separated, coordinate-targeted regime; the controlled benchmark
supplies the null, boundary, shared-majority, and strength sweeps.

The prespecified injected-coordinate median-minus-mean effects were
$-.0387$ to $-.0124$, with every family-four Bonferroni interval below zero.
\par\noindent Targetwise effects were Pop $-.0124$ (CI $[-.0139,-.0109]$), Rock $-.0223$
(CI $[-.0236,-.0208]$), Jazz $-.0333$ (CI $[-.0351,-.0316]$), and Country
$-.0387$ (CI $[-.0405,-.0369]$). The descriptive diagnostics also favored the
median: behavior-anchor top-2 preservation increased by $.0575$--$.2425$ and
anchor $\ell_1$ discrepancy decreased by $.0704$--$.1195$, with ordinary 95\%
intervals excluding zero. This establishes attenuation relative to the
constructed anchor, not latent-preference recovery.

%% file: generated/wsdm27_revision_v3/main_synthetic_confirmation.tex
\begin{tabular}{@{}lrrrrr@{}}
\toprule
DGP & Median & RPCA & $\Delta$ & 95\% CI & Family-2 CI \\
\midrule
Add. & 0.115 & 0.142 & -0.0277 & [-0.0364, -0.0189] & [-0.0379, -0.0175] \\
Logit. & 0.050 & 0.058 & -0.0079 & [-0.0110, -0.0047] & [-0.0116, -0.0042] \\
\bottomrule
\end{tabular}

%% file: generated/wsdm27_revision_v3/main_synthetic_compatible_sets.tex
\begin{tabular}{@{}lrrrrr@{}}
\toprule
$b$ & Declared $(r,\epsilon)$ & Coverage & Mean diam. & Empty & Decision/false (\%) \\
\midrule
0.00 & $(2,0.010)$ & 100.00\% & 0.00997 & 0.00\% & 95.40/0.00 \\
0.35 & $(2,0.010)$ & 100.00\% & 0.01349 & 0.00\% & 97.05/0.00 \\
0.00 & $(2,0.005)$ & 12.49\% & 0.00377 & 50.36\% & 48.85/0.36 \\
0.35 & $(2,0.005)$ & 23.06\% & 0.00421 & 13.62\% & 85.98/0.00 \\
0.00 & $(1,0.010)$ & 0.00\% & -- & 100.00\% & 0.00/0.00 \\
0.35 & $(1,0.010)$ & 0.00\% & -- & 100.00\% & 0.00/0.00 \\
\bottomrule
\end{tabular}

%% file: generated/wsdm27_revision_v3/main_rq1_rq2_absolute_effects.tex
\begin{tabular}{llrrrrrrrr}
\toprule
Target & Metric & Behavior & Median & Rel. red. & Effect & 95\% CI & Bonf. CI & $n$ & W/T/L \\
\midrule
Pop & DNAE & 0.1396 & 0.1291 & 7.6\% & -0.0106 & [-0.01293, -0.00818] & [-0.01383, -0.00736] & 1000 & 633/0/367 \\
Pop & CSE & 0.3413 & 0.3068 & 10.1\% & -0.0344 & [-0.04623, -0.02204] & [-0.05126, -0.01667] & 1000 & 376/404/220 \\
Rock & DNAE & 0.1722 & 0.1709 & 0.7\% & -0.0013 & [-0.00431, 0.00175] & [-0.00552, 0.00296] & 1000 & 534/1/465 \\
Rock & CSE & 0.4152 & 0.3611 & 13.0\% & -0.0541 & [-0.06770, -0.04014] & [-0.07325, -0.03494] & 1000 & 408/354/238 \\
Jazz & DNAE & 0.1521 & 0.1493 & 1.8\% & -0.0028 & [-0.00558, 0.00003] & [-0.00661, 0.00109] & 1000 & 527/5/468 \\
Jazz & CSE & 0.3560 & 0.3199 & 10.1\% & -0.0361 & [-0.04831, -0.02368] & [-0.05356, -0.01882] & 1000 & 329/472/199 \\
Country & DNAE & 0.1477 & 0.1436 & 2.8\% & -0.0041 & [-0.00651, -0.00164] & [-0.00741, -0.00061] & 1000 & 552/4/444 \\
Country & CSE & 0.3263 & 0.3052 & 6.5\% & -0.0211 & [-0.03236, -0.00979] & [-0.03693, -0.00529] & 1000 & 320/462/218 \\
\bottomrule
\end{tabular}

%% file: generated/wsdm27_revision_v3/appendix_amazon_point_method_inventory.tex
\begin{tabular}{@{}lrrrrrrrr@{}}
\toprule
Method & \multicolumn{2}{c}{Pop} & \multicolumn{2}{c}{Rock} & \multicolumn{2}{c}{Jazz} & \multicolumn{2}{c}{Country} \\
\cmidrule(lr){2-3}\cmidrule(lr){4-5}\cmidrule(lr){6-7}\cmidrule(l){8-9}
 & DNAE & CSE & DNAE & CSE & DNAE & CSE & DNAE & CSE \\
\midrule
Behavior only & 0.1396 & 0.3413 & 0.1722 & 0.4152 & 0.1521 & 0.3560 & 0.1477 & 0.3263 \\
Qwen only & 0.1345 & 0.3476 & 0.1724 & 0.3511 & 0.1582 & 0.3555 & 0.1409 & 0.2892 \\
Llama only & 0.1654 & 0.3564 & 0.1941 & 0.4379 & 0.1648 & 0.3651 & 0.1762 & 0.3891 \\
Rating only & 0.1336 & 0.3027 & 0.1785 & 0.4303 & 0.1495 & 0.3456 & 0.1441 & 0.3009 \\
$K=3$ mean (unscaled) & 0.1267 & 0.2971 & 0.1674 & 0.3455 & 0.1488 & 0.3174 & 0.1426 & 0.3145 \\
$K=3$ median (unscaled; primary) & 0.1291 & 0.3068 & 0.1709 & 0.3611 & 0.1493 & 0.3199 & 0.1436 & 0.3052 \\
Behavior only (calibrated) & 0.1396 & 0.3413 & 0.1722 & 0.4152 & 0.1521 & 0.3560 & 0.1477 & 0.3263 \\
$K=3$ median (calibrated) & 0.1292 & 0.3068 & 0.1722 & 0.3611 & 0.1488 & 0.3199 & 0.1429 & 0.3052 \\
Label-first $K=3$ majority & 0.1297 & 0.3020 & 0.1770 & 0.3739 & 0.1525 & 0.3488 & 0.1508 & 0.3335 \\
$K=4$ median (+rating; descriptive) & 0.1253 & 0.2879 & 0.1652 & 0.3658 & 0.1439 & 0.3139 & 0.1383 & 0.2864 \\
\bottomrule
\end{tabular}

%% file: generated/wsdm27_revision_v3/main_rq3_compatible_sets_exact.tex
\begin{tabular}{lrrrrrrrr}
\toprule
Target & $r$ & Covered & Exact 95\% CI & Empty & Sign & Abstain & Elig. sign/coords & Discord./elig. sign \\
\midrule
Pop & 0 & 961/1000 & [0.947, 0.972] & 0.00\% & 0.53\% & 99.47\% & 34/5270 & 8/34 \\
Pop & 1 & 1000/1000 & [0.996, 1.000] & 0.00\% & 0.00\% & 100.00\% & 0/5270 & -- \\
Rock & 0 & 949/1000 & [0.933, 0.962] & 0.06\% & 0.60\% & 99.34\% & 34/5231 & 22/34 \\
Rock & 1 & 1000/1000 & [0.996, 1.000] & 0.00\% & 0.00\% & 100.00\% & 0/5231 & -- \\
Jazz & 0 & 935/1000 & [0.918, 0.949] & 0.04\% & 1.41\% & 98.54\% & 95/5310 & 25/95 \\
Jazz & 1 & 990/1000 & [0.982, 0.995] & 0.00\% & 0.16\% & 99.84\% & 11/5310 & 1/11 \\
Country & 0 & 955/1000 & [0.940, 0.967] & 0.00\% & 0.69\% & 99.31\% & 46/5258 & 24/46 \\
Country & 1 & 1000/1000 & [0.996, 1.000] & 0.00\% & 0.00\% & 100.00\% & 0/5258 & -- \\
\bottomrule
\end{tabular}

%% file: sections/07_conclusion.tex
\newpage
\section{Conclusion}

We establish a sharp $50\%$ boundary: a strict clean majority identifies
support and direction, and known positive scales add magnitude. Exact compatible
sets formalize partial identification. Controlled experiments map the regimes,
confirm the median--RPCA contrast, and yield nontrivial, zero-false-sign
decisions under correct budgets. On Amazon, median aggregation improves all four
CSE targets and the perturbation favors its focal metric; these results concern
fitted surfaces. The framework tells auditors when evidence supports a point, a
set, or remeasurement.

%% file: sections/08_ethical_considerations.tex
\section{Ethical Considerations}

\paragraph{Data and privacy.}
Public reviews and behavioral records can still support intrusive profiling or
identity linkage. The empirical study reports aggregate statistics
and does not release user identifiers, raw reviews, user-level profiles, raw
LLM responses, or record-level attributions. Public availability should not be
interpreted as consent for arbitrary downstream profiling.

\paragraph{Bias, validity, and potential misuse.}
Amazon users, categories, and platform mechanisms are not representative of all
people or preference settings. Behavioral, rating, and LLM proxies can share
platform, selection, and social biases, so agreement does not establish
construct validity. The deterministic zero mapping of parsed-empty LLM profiles
may induce differential measurement error and should be audited across relevant
user groups before reuse. Incorrect preference explanations could enable
manipulative targeting, discriminatory recommendations, or unjustified
inferences about sensitive traits. The method should not be used as sole
evidence in credit, employment, health, education, or other high-stakes
decisions.

\paragraph{Safeguards and uncertainty.}
The strict-majority, cleanliness, scale, reference-distribution, and causal
assumptions must be justified externally. When they are not, the appropriate
outputs are observable diagnostics, compatible-target sets, or abstention rather than
latent-magnitude claims. Recommended safeguards include independent validation,
aggregate-only data release, access controls for private caches, monitoring for
shared proxy bias, human oversight, and an explicit prohibition on
high-stakes automated use.